\documentclass{INTERSPEECH2024}
\usepackage{tikz}
\usepackage{graphicx}
\usepackage{subfig}
\usepackage{floatrow}
\usepackage{multirow}

\usepackage{multirow}
\usepackage{svg}
\usepackage{dirtytalk}
\usepackage{hyperref}
\usetikzlibrary{arrows,positioning,arrows.meta, calc}
\usepackage{ifthen}
\definecolor{darkgreen}{RGB}{11, 153, 56}

\definecolor{extradarkgreen}{RGB}{22, 199, 78}
\usepackage{mathtools}
\usepackage{nicefrac}
\newcommand{\floor}[1]{\floor {#1} \floor}

\interspeechcameraready 

\title{How Much Context Does My Attention-Based ASR System Need?}
\name{Robert}{Flynn}
\name{Anton}{Ragni}

\address{Department of Computer Science, The University of Sheffield, United Kingdom}
\email{\{rjflynn2, a.ragni\}@sheffield.ac.uk}
\keywords{speech recognition, long-context, self-attention}
\begin{document}

\maketitle
 
\begin{abstract}
For the task of speech recognition, the use of more than 30 seconds of acoustic context during training is uncommon and under-investigated in literature. In this work, we conduct an empirical study on the effect of scaling the sequence length used to train/evaluate (dense-attention-based) acoustic models on speech recognition performance. For these experiments, a dataset of roughly 100,000 pseudo-labelled Spotify podcasts is used, with context lengths of 5 seconds to 1 hour being explored. Zero-shot evaluations are presented on the long-format datasets: Earnings-22, Tedlium and Rev16. Results demonstrate a benefit from training with up to 21.8 minutes of acoustic context, showing up to a 14.5\% relative improvement from a baseline trained with 10 seconds of context. We find that the model's width/depth, positional encoding scheme and number of attention heads impact its ability to use longer contexts.

\end{abstract}

\section{Introduction}
Performance on sequence-based tasks has seen a consistent benefit from the introduction of methods that enable the modelling of longer-range dependencies \cite{hochreiter1997longlstm, vaswani2017attention}. The transformer architecture \cite{vaswani2017attention} is a distinct example of this, demonstrating benefits from training on sequences of 1000s of tokens for language modelling \cite{press-etal-2021-shortformer}, and enabling a form of implicit meta-learning known as in-context learning \cite{brown2020language, olsson2022context}. However, for the task of automatic speech recognition (ASR), there is limited work exploring the effect of attending over longer acoustic sequences. In part, this may be due to the format of many academic datasets, which are typically provided as a series of short (typically 1-20s) utterances. This therefore hurts the development of methods that aim to utilise larger amounts of context or learn to segment a recording in a purely end-to-end fashion.

Previous work on utilising cross-utterance acoustic context still deals with fairly short sequences of 20-30s \cite{hori2021advanced, lu2021input, whisperradford2023robust, chiu2019comparison}. Other work employs local attention \cite{rekesh2023fastconformer, koluguri2023investigating}, typically with a window duration of 10 seconds, limiting the model's ability to use the full context.  The often stated reason for limiting the context window used by self-attention-based models is their quadratic complexity with respect to the sequence length. However, it is also not clear whether this current modelling paradigm is capable of utilising truly long sequences of minutes or hours in duration. For instance, for the task of language modelling, \cite{press-etal-2021-shortformer} finds that transformers struggle to gain any benefit from sequences longer than 1024 when trained on their target dataset. Similar findings are reported in \cite{sun2021long}, for language models trained with local attention.

%While long-context acoustic models (AMs) are fairly under-investigated, there is ample work \cite{sun2021transformercrossutt, flynn2023leveraging, chiu2021crossuttbert} on utilising cross-utterance context within the language modelling component of ASR systems. However, we argue this is not optimal, as an assumption of independence between utterances is still made at the acoustic-level. Consequently, the long-context language model may not recover information lost by the utterance-level AM, and the system is only able to adapt to linguistic aspects of the data based on the context.

As such, in this work, we investigate the benefit of scaling the context/sequence length used during the training and evaluation of dense-attention-based acoustic models (AMs), to better understand the capabilities of current methods. To assess this, models of varying context lengths are trained on a large collection of podcasts and evaluated on a set of long-format datasets. We also vary the positional encodings, number of layers and attention heads in order to understand the impact these factors may have on the model's ability to use longer contexts. A breakdown of our contributions is given as follows:
\begin{enumerate}
    \item We demonstrate ($\S$ \ref{sec:howmuchcontext}) a benefit from training with up to 21.8 minutes of acoustic context, and successfully train an AM with 1 hour of context without degradation, a magnitude larger than what is used in literature.
    \item We show ($\S$ \ref{sec:robusttonoise}) that models trained on longer contexts are more robust to domain shifts and provide some insight as to why this may happen.
    \item We asses ($\S$ \ref{sec:whatposmethod}) multiple positional encoding methods and find that rotary encodings lead to increasingly better performance as the context size is scaled when compared to sinusoidal encodings.

    \item We demonstrate ($\S$ \ref{sec:whatmodelsize}) that the models must be sufficiently deep and wide in order to benefit from longer contexts. Additionally, we find that having more attention heads, with a smaller head dimension is helpful at shorter sequence lengths, but harmful on longer sequences.
    % \item Finally, despite only training for a single epoch, we show that the use of data augmentation is still helpful and can encourage the model to learn to use more long-range context.
\end{enumerate}

%As such, in this work, we investigate the benefit of extending the context length of both the AM and language model (LM) components of transformer based speech recognition systems. A breakdown of our contributions is given as follows: \textbf{1.} An investigation is conducted on the effect of training/evaluation context length on speech recognition, with results demonstrating an optimal context length of around 80s. \textbf{2.} We demonstrate training of dense-attention based AMs with maximum context lengths of up to \textbf{1 hour} through the application of a sequence length warmup and various efficiency adaptions from prior work. \textbf{3.} An overlapping decoding scheme is introduced to reduce context fragmentation, and the optimal amount of overlap to use is investigated.

%The rest of this work is ordered as follows: Section \ref{longcontextadaptions} details the training and evaluation adaptions that were made in order to fairly investigate and compare a range of context sizes for both the AM and LM components of the ASR system. Section \ref{expdetails} overviews the experimental details. Section \ref{results} presents and analyses the results, with the conclusion given in section \ref{conclusion}. 

Finally, we release all trained model checkpoints and code.\footnote{\url{www.github.com/robflynnyh/long-context-asr}}

\vspace{-0.5em}
\section{Modifications for Long-Context ASR}
\label{longcontextadaptions}
\subsection{Architecture}
\begin{table}[hbt]
    \footnotesize
    \centering\
    \begin{tabular}{l|c|c}
        Subsampling Scheme &  Duration (min) &  Speed $ \uparrow$ (frames/s)  \\ \hline
        Conformer    &    9 / 23 &    30,200  / 57,300     \\
        FastConformer &   18 / 70 &  57,650 / 80,900         \\
    \end{tabular}
    \caption{Maximum context possible on 1 A100 during \textbf{training} with batch size of 1. Without/With Flash Attention.}
    \label{tab:conformerfastconformerspeed}
\end{table}

\noindent Conformer-based AMs \cite{gulati2020conformer}, trained with connectionist temporal classification (CTC) \cite{graves2006connectionist}, are used as the basis for this investigation.
These architectures typically utilise some form of subsampling. Recent work \cite{rekesh2023fastconformer} explores increasing the level of subsampling from $4\times$ to $8\times$, as a simple method of decreasing the sequence length and hence reducing the compute and memory complexity of the model. Additionally, the standard convolution blocks in the subsampling module are replaced with depthwise separable convolutions with a smaller feature dimension than the rest of the model. This configuration is referred to as FastConformer, with results demonstrating favourable accuracy-efficiency trade-offs compared to the Conformer. As shown in table \ref{tab:conformerfastconformerspeed}, when paired with flash-attention \cite{dao2022flashattention} (an efficient algorithm for computing attention on the GPU without approximations), this modification makes it possible to train on recordings longer than an hour on an 80 GB A100 GPU.  

\vspace{-0.5em}

\subsection{Moving window decoding}
Typically, in ASR, a long-format dialogue is segmented into a set of utterances based on silences, and these utterances are transcribed independently. As a consequence, frames near the start and end of an utterance have a fragmented context, which can harm performance \cite{press-etal-2021-shortformer}. As the sequence length used by the model is increased, the number of positions where the context is fragmented would therefore be reduced. This can result in an illusion of benefit from many long-context methods if evaluated naively, as the model will show better performance due to a reduction in context fragmentation, without necessarily using any long-range context \cite{press-etal-2021-shortformer}.

% \begin{figure}
%     \centering
%     \includegraphics[width=6.5cm]{figures/overlaps.pdf}
%     \caption{Depiction of overlapping window inference.}
%     \label{fig:overlapping_inference}
% \end{figure}

To avoid fragmenting the context and therefore enable fairer comparison between different context lengths, recordings are processed using a moving window decoding scheme. Specifically, the input is processed in windows $W \ni (w_0... w_N)$ and for a given stride $s$ the starting position of the $i^{th}$ window $w_i$ is given by: $i\cdot s$. When the stride is smaller than the sequence length, this results in multiple predictions at various frames, which are averaged to obtain the final predictions.

\vspace{-0.5em}
\subsection{Sequence length warmup} 
Training attention-based models on long sequences can result in instability in early training due to gradient variance \cite{li2022stabilityefficiency}. We find this gradient variance to be destructive for the AM when the context is greater than 40s, with these models often failing to train. To avoid this, a sequence length warmup \cite{press-etal-2021-shortformer, li2022stabilityefficiency} is used, where the sequence length is gradually increased throughout training. For this method several hyperparameters are employed, namely: a minimum sequence length $s_0$ that is used at the start of training, which is then doubled every $n$ recordings/steps $r$, until a maximum sequence length $s_m$ is reached. Hence, the sequence length at a given recording $s_r$ is shown as follows: $s_r = \min(s_0 + s_0 \cdot 2 ^{\lfloor \nicefrac{r}{n} \rfloor}, s_m)$

\vspace{-0.5em}
\subsection{Positional encoding}
\label{sec:pos_enc_explanation}
Transformer-type models generally employ some form of positional encoding method \cite{vaswani2017attention}. As the method of encoding positions may be crucial for informing token-token interactions over long distances, we investigate several approaches for this work, which are as follows: sinusoidal encodings \cite{vaswani2017attention}, that employ sine and cosine functions which are added to the input to encode the absolute position; no positional encodings, which rely on the convolutional layers in the Conformer to encode position; rotary encodings \cite{su2021roformer, li2021conformer}, which encode absolute and relative position by applying a rotation matrix to the keys and queries prior to self-attention. 

Rotary encodings employ a hyperparameter $\theta$ which acts as a base period, controlling the amount of rotation between tokens, with smaller values of $\theta$ biasing the model more towards nearby tokens. While $\theta$ is commonly set to a value of 10K, increasing $\theta$ has shown to be a beneficial modification when working with longer sequences \cite{roziere2023code}. Hence, in this work, we also investigate other values of $\theta$ and report on a setting where $\theta=1.5$M.

\vspace{-0.5em}
\section{Experimental Configuration}
\label{expdetails}
\subsection{Data}

%\subsubsection{Acoustic Model}
% \begin{table}[]
%     \centering
%     \begin{tabular}{c|c|c|c}
%         Corpus & Files & Hours  & Usage \\\hline
%         Spotify Podcasts  &  105,360  & 58,000 & Train   \\
%         Tedlium & 8 / 11 & 1.6 / 2.6 & Dev / Test \\
%         Earnings-22 & 6 / 6 &  5.5 / 5.6 & Dev / Test \\
%     \end{tabular}
%     \caption{Corpus statistics for AM data}
%     \label{tab:corpus_stats}
% \end{table}

As investigating long-context models may require larger amounts of long-format data than typical ASR datasets provide, the collection of Spotify podcasts provided in \cite{spotify-clifton-etal-2020-100000} is selected for the AMs training data. Podcasts in this dataset last on average 33 minutes, with many going over one hour. In total, this amounts to 58K hours of training data. This data is not human-labeled and instead is provided with pseudo-labels produced using Google's speech API. Tedlium \cite{hernandez2018ted}, Earnings-22 \cite{del2022earnings} and Rev16 \cite{whisperradford2023robust} are used as evaluation datasets, which were selected due to their long-format. 

Tedlium is composed of single-speaker TED talks lasting around 14 minutes in duration. Tedlium's dev and test sets total 1.6 and 2.6 hours respectively. As Tedlium contains segments of untranscribed speech such as adverts, these portions of the spectrogram are set to zero for moving window decoding. Earnings-22 consists of earning report meetings lasting up to several hours with multiple speakers and a diverse range of accented speech. We use the entire dataset for evaluation\footnote{As opposed to the smaller test set used in \cite{gandhi2022esb, open-asr-leaderboard}}, which totals 125 meetings and 119 hours. Rev16 is composed of podcasts and can be seen as our in-domain test set, we use the 16 recordings detailed in \cite{whisperradford2023robust}, totalling 16.2 hours.

All audio data is converted to 16khz, and 80-band Mel spectrograms (window length of 400 and hop length of 160) are used to train the model. 
Mean and standard deviation statistics across each recording are used for spectrogram normalisation.
For text tokenization, the sentencepiece tokenizer is used with the \say{nmt\_nfkc\_cf} normalisation rule. As the model is not able to adapt to dataset specific transcription styles, normalisation is applied to any model outputs and the reference transcript. The text normaliser from Whisper \cite{whisperradford2023robust} is used.

\vspace{-0.5em}
\subsection{Model configuration}
The AM uses the FastConformer \cite{rekesh2023fastconformer} subsampling configuration with 8x downsampling using depthwise separable convolutions with a hidden dimension of 256, followed by $N$ Conformer layers \cite{gulati2020conformer}. The model is trained using SC-CTC \cite{nozaki2021relaxingselfconditioned}, without intermediate losses. Batch normalisation \cite{ioffe2015batch} is replaced with batch renormalisation \cite{ioffe2017batchrenorm}, and convolutional modules use a kernel size of 9. The flash attention algorithm \cite{dao2022flashattention} is used to compute attention. All models use a vocabulary composed of 4095 BPE tokens learnt from the Spotify corpus with an additional blank token. 

For our primary investigations, a model with 6 attention heads, 6 layers, and a hidden dimension of 768 is used (totalling around 90 million parameters). Rotary position encoding with $\theta = 1.5$M is used for all experiments excluding those in $\S$ \ref{sec:whatposmethod} where position encoding is varied. For the model scaling experiments ($\S$ \ref{sec:whatmodelsize}) we experiment with various hidden dimensions, number of layers and number of heads. The largest model configuration used 3 layers, 2048 hidden dimension and 16 heads for a total of 315M parameters.

\vspace{-0.5em}
\subsection{Training configuration}
To ensure all context sizes receive the same number of optimization steps, the total duration of each batch is kept fixed at 1 hour of audio. 
As the training data is provided with word-level timesteps, the podcasts can be chunked into inputs of arbitrary length and the text corresponding to each chunk can be retrieved for training. Context sizes in seconds of the following lengths are used for the investigations: 10, 20, 41, 82, 164, 328, 655, 1311, 2621 and 3600.

The Madgrad optimizer \cite{defazio2022adaptivitymadgrad} is used for all training runs with gradient clipping and a learning rate warmup followed by a cosine annealing schedule. A learning rate of around $3e-3$ is used for most experiments. The models are trained with a sequence length warmup, 5s is used as the initial sequence length $s_0$, which is doubled every 5K recordings. All models are trained for one epoch on an 80GB A100 GPU, and 321 models are trained in total. Training takes around 15–24 hours, for context lengths below 20 minutes, and 65 hours for the maximum context length of 1 hour. Due to the use of Flash Attention \cite{dao2022flashattention} and the fixed batch duration, memory consumption is constant across all sequence lengths.

%For experiments using data augmentation SpecAugment \cite{park2019specaugment} is used. For frequency masking, a mask width of up to 32 with 3 masks is used. For time masking 5 masks are used, with a combined with of 5\% of the sequence length. 

% \begin{figure}[hbt]
%     \centering
%     \includegraphics[width=7cm]{figures/augmentation.pdf}
%     \caption{WERs on Earnings-22 while varying sequence length and data augmentation and using Fourier positional encodings.}
%     \label{fig:enter-label}
% \end{figure}
\vspace{-0.5em}
\section{Experimental Results}
All experiments use three repeats, with different random seeds, and mean word error rates (WERs) are reported in the figures. Results at each sequence length are from \textbf{separate models} that are both trained and evaluated at that length. The moving window decoding scheme is used for all evaluations with a stride equal to $12.5\%$ of the sequence length. Using a stride smaller than this did not affect the results. The following subsections discuss our various findings.

% \textbf{Remaining experiments to run:}

% \begin{itemize}
%     \item different head sizes
%     \item evaluation where 1 hour model is evaluated at different sequence lengths, but with zero padding so sequence length passed to the model is constant, to show model is not better in general
% \end{itemize}
\vspace{-0.5em}
\subsection{How much context is useful?}
\label{sec:howmuchcontext}
\begin{figure}[hbt]
    \centering
    \includegraphics[width=6cm]{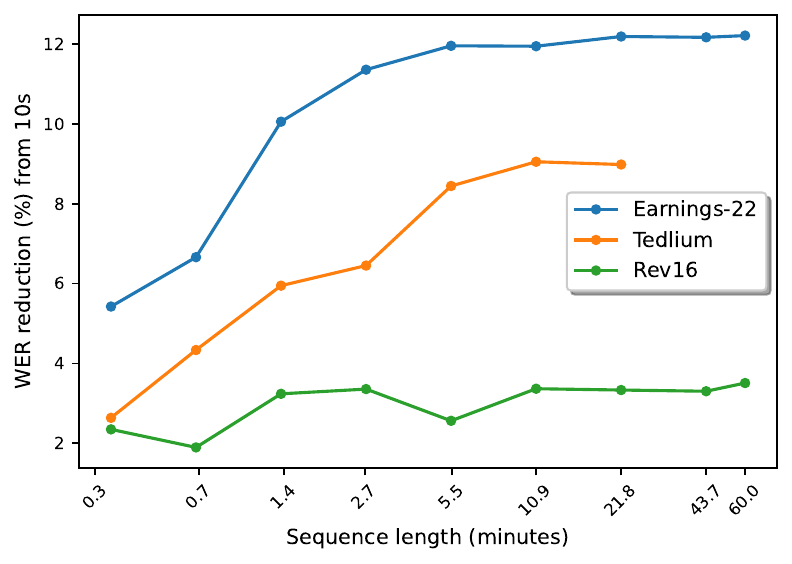}
    \caption{WER reduction from a sequence length of 10s}
    \label{fig:werr_across_datasets}
\end{figure}
\noindent A comparison of word error rate reductions (WERRs) from a baseline with 10s of context, as the sequence length is increased, for each of the evaluation datasets is given in figure \ref{fig:werr_across_datasets}. This 10s baseline has a WER of 27.7\%, 6.8\% and 15.0\% on Earnings-22, Tedlium and Rev16 respectively - these results are in a similar range to the Whisper tiny model (table D.4) \cite{whisperradford2023robust}.
As shown, longer sequence lengths are the most beneficial on Earnings-22, our most challenging and noisy dataset. On this data, the models show the largest improvement as the context size is scaled, with up to a 12.2\% WERR from the 10s baseline. Training and evaluating with up to 21.8 minutes of context was beneficial on Earnings-22. Note that due to variance in the repeats the WER at 20 minutes was not significantly lower than at 10 minutes ($p=0.592$), however, the WER at 43.6 minutes ($p=0.024$) and 1 hour ($p=0.009$) was significantly lower than at 10 minutes. For Tedlium the model continues improving up to 5.5 minutes of context, no recordings longer than 21.8 minutes were present in this dataset. For both Earnings-22 and Tedlium, the majority of the WERR is reached at 5.5 minutes of context, hence this may be the most practical sequence length to train/evaluate at, in terms of accuracy/efficiency based on our results. 

%($p=0.015$) for both

On Rev16, our in-domain dataset, we do not see any significant benefit to training on sequences of longer than 20s. We suspect this behaviour is due to the similarity of Rev16 to our large training dataset, allowing the model to rely more on the information stored in the weights than the context. A similar trend is displayed when evaluating the WER/loss on a small subset of the training data.  This then begs the question as to why the model learns to use the long-range context when it is able to amortize the necessary information in the weights during training at a shorter context. 
%From inspecting the predictions on Rev16 we find that the longer-context models are more confident in their predictions. Additionally, while the WER does not consistently decease beyond 20s on the training data the loss decreases up until 2.7 minutes of context, however this also plateus beyond this point. 
We hypothesise that providing more context acts as an inductive bias, causing the longer-context models to allocate some of their parameters towards solutions that use this extra context. While this does not necessarily result in better performance on the training domain, these solutions may be more general, and therefore more robust to changes in the domain.

% model is incentivized during training to use the long-range context if it is able to amortize this information in the weights. When inspecting the model outputs, the predictions from the longer-context models show higher confidence/probability. 

%Additonally, when the models are evaluated on a subset of the training data, they show a similar tredn

% We find that the model does in-fact continue reducing its loss on Rev16 up until X minutes of context, but these improvements are not reflected in the WER. Hence, at longer contexts, the model is becoming more confident for words which are already correctly transcribed at shorter contexts. We believe this behaviour causes the long-context model to become robust to domain shifts at test time.

\label{sec:robusttonoise}
\begin{figure}[hbt]
    \centering
    \includegraphics[width=6cm]{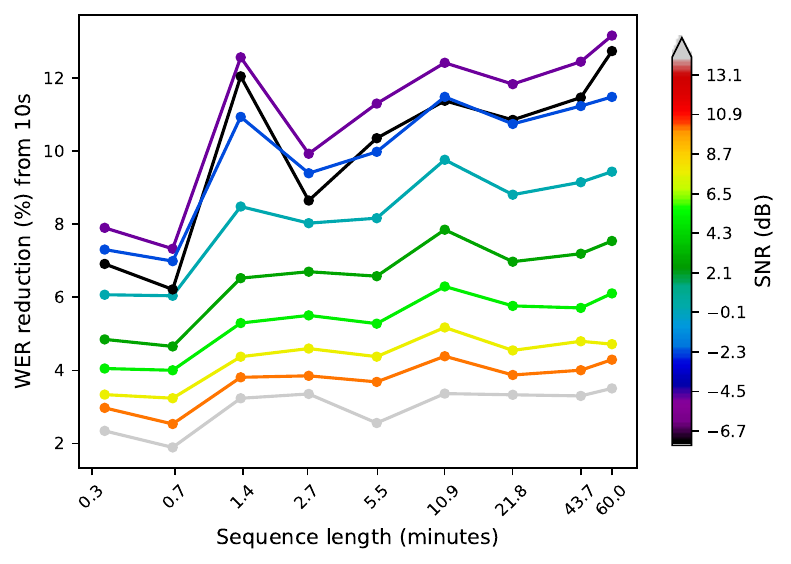}
    \caption{WER reduction from 10s baseline on Rev16 with varying amounts of background music}
    \label{fig:robusttonoise}
\end{figure}

\noindent As context lengths longer than 20s were not beneficial on Rev16 we experiment with artificially adding noise (music) at various signal-to-noise ratios (SNRs) to see if the longer contexts become useful after this domain shift. Results are presented in figure \ref{fig:robusttonoise}. The results demonstrate that models trained on a longer context show a greater WERR, as the SNR is increased, and are therefore more robust to this form of domain shift. We found a similar trend when using other forms of noise. Additionally, this helps validate our hypothesis that longer contexts are less beneficial for data that is similar to the training data. The large increase seen at 80s at low SNRs is due to random variation in the different model's zero-shot robustness to very large amounts of noise. While the results show variation, the general trend suggests that the WERR increases up to 1 hour of context for lower SNRs. Unlike in figure \ref{fig:werr_across_datasets}, the majority of this benefit is attained with 20s of context, hence the models are primarily using the local context to adapt to the noise. This suggests that other forms of adaptation (i.e. linguistic) may be taking place for Earnings-22 and Tedlium.

\vspace{-0.5em}
\subsection{Impact of positional encoding method}

\label{sec:whatposmethod}
\begin{figure}[hbt]
    \centering
    \includegraphics[width=6cm]{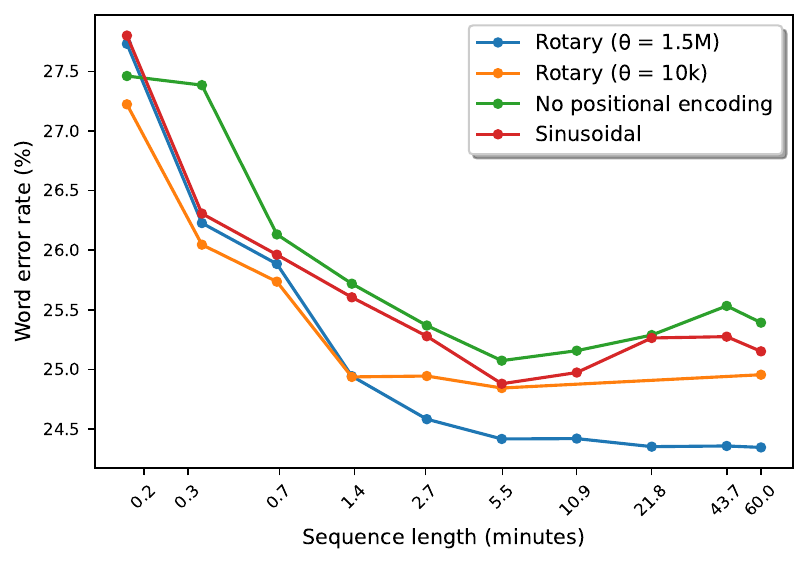}
    \caption{WER at various sequence length for different positional encoding methods on Earnings-22}
    \label{fig:posmethod}
\end{figure}
\noindent Figure \ref{fig:posmethod} presents our results on Earnings-22 when training at different sequence lengths with the positional encoding schemes described in \ref{sec:pos_enc_explanation}. 
Interestingly, we find that there is little difference between each of the positional encoding schemes when using a sequence length of 10s. As the Conformer architecture uses convolutions during subsampling and at every layer, this may be sufficient to encode positional information at shorter sequence lengths. However, for sequence lengths greater than 80s Rotary ($\theta = 1.5$M) shows better performance, with a $3.2\%$ WERR compared to Sinusoidal at a sequence length of 1 hour. Rotary ($\theta = 1.5$M) was the only encoding method that was able to benefit from a sequence length greater than 328s. When using the lower $\theta$ value of 10k, the performance plateaued at 80s of context, showing that the distance bias prevented the model from using the longer contexts. Increasing $\theta$ further did not allow the model to benefit from context lengths greater than 21.8 minutes. These results demonstrate that the positional encoding method is important to consider when training longer context models, and may be a promising direction for future research for trying to leverage even longer contexts. 

\vspace{-0.5em}
\subsection{Impact of model size}

\label{sec:whatmodelsize}
\vspace{-1em}
\begin{figure}[hbt]

    \centering
    \includegraphics[width=6cm]{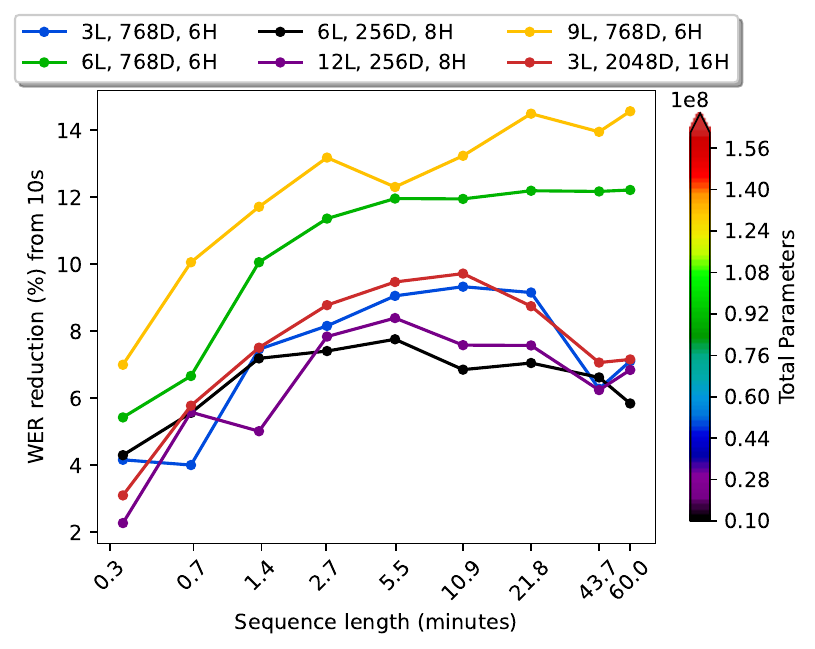}
    \vspace{-0.5em}
    \caption{WER reduction from a sequence length of 10s for various model sizes on Earnings-22}
    \label{fig:modelsizes}
\end{figure}

\noindent To investigate whether model size (\textbf{L}ayers, Hidden \textbf{D}imension, Attention \textbf{H}eads) affects the model's ability to use longer contexts we experiment with various configurations which are presented in figure \ref{fig:modelsizes}. 
The results show that the models with less than 90 million parameters showed worse performance on Earnings-22 when trained/evaluated on sequences longer than 5-10 minutes, and show lower WERR across all sequence lengths. Similar behaviour is shown for the largest model with 315M parameters and 3 layers. More investigation is needed to understand why this degeneration occurs, although it is clear that it is crucial to have a sufficiently deep and wide model when working with longer contexts. Interestingly, none of the model configurations display this degeneration on Rev16, where the WER remains fairly constant beyond 20s of context.

The largest WERR of 14.5\% was seen from the 9L model with 130M parameters. While this model shows a larger WERR than the 90M parameter, it is still not able to benefit from a sequence length larger than 21.8 minutes. This suggests that there is a bottleneck other than scale, however, deeper and wider models would ideally be investigated given greater resources.

\begin{figure}[hbt]
    \centering
    \includegraphics[width=6cm]{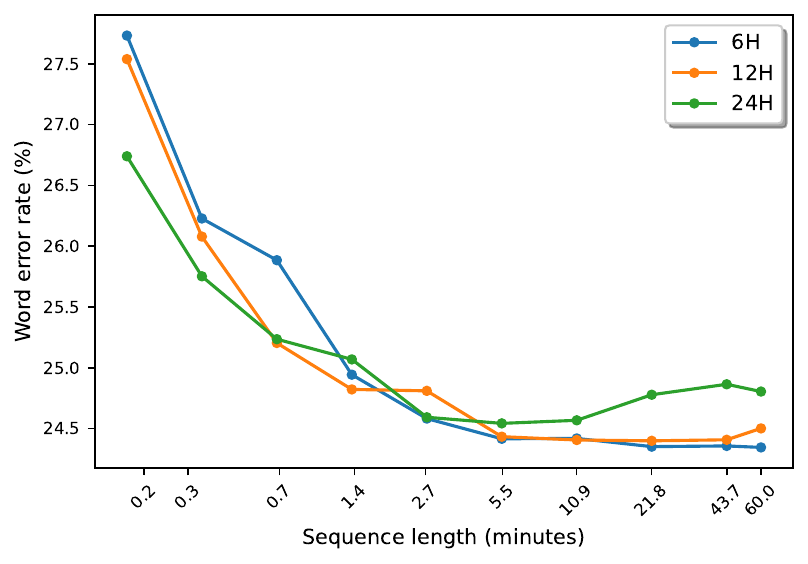}
    \caption{WER on Earnings-22 when varying the number of attention heads (H)}
    \label{fig:headsizes}
\end{figure}

\noindent Figure \ref{fig:headsizes} presents experiments where the number of attention heads and the sequence length are varied for a model with 9 layers and a hidden dimension of 768. Here, we find that configurations with more heads (and a smaller head dimension) perform better at shorter sequence lengths and worse at longer sequence lengths. Both the 24 and 12h configurations begin to increase in WER at longer sequence lengths, with the 24h configuration showing the most degradation. The 6-head model is the only configuration that is able to benefit from the full 21.8 minutes of context.

As noted in \cite{vaswani2017attention, shazeer2020talking}, increasing the number of attention heads becomes counterproductive when the head dimension decreases below 32. \cite{shazeer2020talking} posits that this is due to the keys and queries becoming \say{so low-dimensional that their dot product can no longer constitute an informative
matching function.} We hypothesise that this effect becomes more severe as the sequence length is increased, requiring a more accurate matching function as the sparsity of relevant information is increased. This may be a similar phenomenon to the degradation seen in figure \ref{fig:modelsizes} at longer sequence lengths for the smaller models.  

\vspace{-0.5em}
\section{Conclusion}
Many use-cases for ASR involve long-format data i.e. meetings or lectures, consequently, there is a demand for models that can utilise the large amount of context information that is present in these formats. To better understand the capabilities of current approaches, this work investigated the impact of altering various architectural components, along with the amount of context, used during the training/evaluation of dense attention-based ASR systems. These results illustrate the approach of training on longer sequences is a simple but effective method of improving model performance when long-format data is available. We find that it is beneficial to train with up to 21.8 minutes of context, a magnitude larger than what is used in literature. Notably, the models trained on longer sequence lengths were more robust to domain shifts at test time, and we provide some insight as to why this behaviour may occur. Architectural components that impact the model's use of very long context were also identified, which may provide directions for improving attention-based AMs. While these experiments are not exhaustive, the results suggest a limit to the amount of context current dense attention-based AMs can benefit from. In future, we plan to investigate other architectures and conduct further interpretability work to better ascertain how the model is using very long-contexts. 

%% SPELL CHECK.

% \begin{figure}[hbt]
%     \centering
%     \includegraphics[width=7cm]{figures/model_sizes_dev.pdf}
%     \caption{WER reduction from a sequence length of 10s for various model sizes on Earnings-22 (dev)}
%     \label{fig:enter-label}
% \end{figure}

%\subsection{Impact of Data Augmentation}

\section{Acknowledgements}
This work was supported by the CDT in Speech and Language Technologies (SLT) and their Applications funded by UKRI [grant number
EP/S023062/1].

\bibliographystyle{IEEEtran}
\bibliography{mybib}

% Generated by IEEEtran.bst, version: 1.13 (2008/09/30)
\begin{thebibliography}{10}
\providecommand{\url}[1]{#1}
\csname url@samestyle\endcsname
\providecommand{\newblock}{\relax}
\providecommand{\bibinfo}[2]{#2}
\providecommand{\BIBentrySTDinterwordspacing}{\spaceskip=0pt\relax}
\providecommand{\BIBentryALTinterwordstretchfactor}{4}
\providecommand{\BIBentryALTinterwordspacing}{\spaceskip=\fontdimen2\font plus
\BIBentryALTinterwordstretchfactor\fontdimen3\font minus \fontdimen4\font\relax}
\providecommand{\BIBforeignlanguage}[2]{{%
\expandafter\ifx\csname l@#1\endcsname\relax
\typeout{** WARNING: IEEEtran.bst: No hyphenation pattern has been}%
\typeout{** loaded for the language `#1'. Using the pattern for}%
\typeout{** the default language instead.}%
\else
\language=\csname l@#1\endcsname
\fi
#2}}
\providecommand{\BIBdecl}{\relax}
\BIBdecl

\bibitem{hochreiter1997longlstm}
S.~Hochreiter and J.~Schmidhuber, ``Long short-term memory,'' \emph{Neural computation}, vol.~9, no.~8, pp. 1735--1780, 1997.

\bibitem{vaswani2017attention}
A.~Vaswani, N.~Shazeer, N.~Parmar, J.~Uszkoreit, L.~Jones, A.~N. Gomez, {\L}.~Kaiser, and I.~Polosukhin, ``Attention is all you need,'' \emph{NeurIPS}, vol.~30, 2017.

\bibitem{press-etal-2021-shortformer}
O.~Press, N.~A. Smith, and M.~Lewis, ``Shortformer: Better language modeling using shorter inputs,'' \emph{arXiv preprint arXiv:2012.15832}, 2020.

\bibitem{brown2020language}
T.~Brown, B.~Mann, N.~Ryder, M.~Subbiah, J.~D. Kaplan, P.~Dhariwal, A.~Neelakantan, P.~Shyam, G.~Sastry, A.~Askell \emph{et~al.}, ``Language models are few-shot learners,'' \emph{Advances in neural information processing systems}, vol.~33, pp. 1877--1901, 2020.

\bibitem{olsson2022context}
C.~Olsson, N.~Elhage, N.~Nanda, N.~Joseph, N.~DasSarma, T.~Henighan, B.~Mann, A.~Askell, Y.~Bai, A.~Chen \emph{et~al.}, ``In-context learning and induction heads,'' \emph{arXiv preprint arXiv:2209.11895}, 2022.

\bibitem{hori2021advanced}
T.~Hori, N.~Moritz, C.~Hori, and J.~L. Roux, ``Advanced long-context end-to-end speech recognition using context-expanded transformers,'' \emph{arXiv preprint arXiv:2104.09426}, 2021.

\bibitem{lu2021input}
Z.~L, Y.~Pan, T.~Doutre, P.~Haghani, L.~Cao, R.~Prabhavalkar, C.~Zhang, and T.~Strohman, ``Input length matters: Improving rnn-t and mwer training for long-form telephony speech recognition,'' \emph{arXiv preprint arXiv:2110.03841}, 2021.

\bibitem{whisperradford2023robust}
A.~Radford, J.~W. Kim, T.~Xu, G.~Brockman, C.~McLeavey, and I.~Sutskever, ``Robust speech recognition via large-scale weak supervision,'' in \emph{ICML}.\hskip 1em plus 0.5em minus 0.4em\relax PMLR, 2023, pp. 28\,492--28\,518.

\bibitem{chiu2019comparison}
C.-C. Chiu, W.~Han, Y.~Zhang, R.~Pang, S.~Kishchenko, P.~Nguyen, A.~Narayanan, H.~Liao, S.~Zhang, A.~Kannan \emph{et~al.}, ``A comparison of end-to-end models for long-form speech recognition,'' in \emph{2019 IEEE automatic speech recognition and understanding workshop (ASRU)}.\hskip 1em plus 0.5em minus 0.4em\relax IEEE, 2019, pp. 889--896.

\bibitem{rekesh2023fastconformer}
D.~Rekesh, S.~Kriman, S.~Majumdar, V.~Noroozi, H.~Huang, O.~Hrinchuk, A.~Kumar, and B.~Ginsburg, ``Fast conformer with linearly scalable attention for efficient speech recognition,'' \emph{arXiv preprint arXiv:2305.05084}, 2023.

\bibitem{koluguri2023investigating}
N.~R. Koluguri, S.~Kriman, G.~Zelenfroind, S.~Majumdar, D.~Rekesh, V.~Noroozi, J.~Balam, and B.~Ginsburg, ``Investigating end-to-end asr architectures for long form audio transcription,'' \emph{arXiv preprint arXiv:2309.09950}, 2023.

\bibitem{sun2021long}
S.~Sun, K.~Krishna, A.~Mattarella-Micke, and M.~Iyyer, ``Do long-range language models actually use long-range context?'' \emph{arXiv preprint arXiv:2109.09115}, 2021.

\bibitem{gulati2020conformer}
A.~Gulati, J.~Qin, C.~Chiu, N.~Parmar, Y.~Zhang, J.~Yu, W.~Han, S.~Wang, Z.~Zhang, Y.~Wu \emph{et~al.}, ``Conformer: Convolution-augmented transformer for speech recognition,'' \emph{arXiv preprint arXiv:2005.08100}, 2020.

\bibitem{graves2006connectionist}
A.~Graves, S.~Fern{\'a}ndez, F.~Gomez, and J.~Schmidhuber, ``Connectionist temporal classification: labelling unsegmented sequence data with recurrent neural networks,'' in \emph{ICML}, 2006.

\bibitem{dao2022flashattention}
T.~Dao, D.~Y. Fu, S.~Ermon, A.~Rudra, and C.~Ré, ``Flashattention: Fast and memory-efficient exact attention with io-awareness,'' 2022.

\bibitem{li2022stabilityefficiency}
C.~Li, M.~Zhang, and Y.~He, ``The stability-efficiency dilemma: Investigating sequence length warmup for training gpt models,'' \emph{NeurIPS}, vol.~35, pp. 26\,736--26\,750, 2022.

\bibitem{su2021roformer}
J.~Su, Y.~Lu, S.~Pan, A.~Murtadha, B.~Wen, and Y.~Liu, ``Roformer: Enhanced transformer with rotary position embedding,'' \emph{arXiv preprint arXiv:2104.09864}, 2021.

\bibitem{li2021conformer}
S.~Li, M.~Xu, and X.-L. Zhang, ``Conformer-based end-to-end speech recognition with rotary position embedding,'' in \emph{2021 Asia-Pacific Signal and Information Processing Association Annual Summit and Conference (APSIPA ASC)}.\hskip 1em plus 0.5em minus 0.4em\relax IEEE, 2021, pp. 443--447.

\bibitem{roziere2023code}
B.~Roziere, J.~Gehring, F.~Gloeckle, S.~Sootla, I.~Gat, X.~E. Tan, Y.~Adi, J.~Liu, T.~Remez, J.~Rapin \emph{et~al.}, ``Code llama: Open foundation models for code,'' \emph{arXiv preprint arXiv:2308.12950}, 2023.

\bibitem{spotify-clifton-etal-2020-100000}
\BIBentryALTinterwordspacing
A.~Clifton, S.~Reddy, Y.~Yu, A.~Pappu, R.~Rezapour, H.~Bonab, M.~Eskevich, G.~Jones, J.~Karlgren, B.~Carterette, and R.~Jones, ``100,000 podcasts: A spoken {E}nglish document corpus,'' in \emph{COLING}.\hskip 1em plus 0.5em minus 0.4em\relax ICCL, Dec. 2020, pp. 5903--5917. [Online]. Available: \url{https://aclanthology.org/2020.coling-main.519}
\BIBentrySTDinterwordspacing

\bibitem{hernandez2018ted}
F.~Hernandez, V.~Nguyen, S.~Ghannay, N.~Tomashenko, and Y.~Esteve, ``Ted-lium 3: Twice as much data and corpus repartition for experiments on speaker adaptation,'' in \emph{SPECOM}.\hskip 1em plus 0.5em minus 0.4em\relax Springer, 2018, pp. 198--208.

\bibitem{del2022earnings}
M.~Del~Rio, P.~Ha, Q.~McNamara, C.~Miller, and S.~Chandra, ``Earnings-22: A practical benchmark for accents in the wild,'' \emph{arXiv preprint arXiv:2203.15591}, 2022.

\bibitem{gandhi2022esb}
S.~Gandhi, P.~Von~Platen, and A.~M. Rush, ``Esb: A benchmark for multi-domain end-to-end speech recognition,'' \emph{arXiv preprint arXiv:2210.13352}, 2022.

\bibitem{open-asr-leaderboard}
V.~Srivastav, S.~Majumdar, N.~Koluguri, A.~Moumen, S.~Gandhi \emph{et~al.}, ``Open automatic speech recognition leaderboard,'' \url{https://huggingface.co/spaces/hf-audio/open_asr_leaderboard}, 2023.

\bibitem{nozaki2021relaxingselfconditioned}
J.~Nozaki and T.~Komatsu, ``Relaxing the conditional independence assumption of ctc-based asr by conditioning on intermediate predictions,'' \emph{arXiv preprint arXiv:2104.02724}, 2021.

\bibitem{ioffe2015batch}
S.~Ioffe and C.~Szegedy, ``Batch normalization: Accelerating deep network training by reducing internal covariate shift,'' in \emph{ICML}.\hskip 1em plus 0.5em minus 0.4em\relax pmlr, 2015, pp. 448--456.

\bibitem{ioffe2017batchrenorm}
S.~Ioffe, ``Batch renormalization: Towards reducing minibatch dependence in batch-normalized models,'' \emph{NeurIPS}, 2017.

\bibitem{defazio2022adaptivitymadgrad}
A.~Defazio and S.~Jelassi, ``Adaptivity without compromise: a momentumized, adaptive, dual averaged gradient method for stochastic optimization,'' \emph{J Mach Learn Res}, vol.~23, 2022.

\bibitem{shazeer2020talking}
N.~Shazeer, Z.~Lan, Y.~Cheng, N.~Ding, and L.~Hou, ``Talking-heads attention,'' \emph{arXiv preprint arXiv:2003.02436}, 2020.

\end{thebibliography}

\end{document}